\pdfoutput=1
\documentclass[11pt]{article}

\usepackage[preprint]{acl}

\usepackage{times}
\usepackage{latexsym}
\usepackage{adjustbox}
\usepackage{multirow}
\usepackage{enumitem}
\usepackage{xcolor}
\usepackage{soul}
\usepackage[T1]{fontenc}
\usepackage[utf8]{inputenc}

\usepackage{microtype}

\usepackage{inconsolata}

\usepackage{graphicx}
\usepackage{booktabs}
\usepackage{amssymb}% http://ctan.org/pkg/amssymb
\usepackage{pifont}% http://ctan.org/pkg/pifont
\title{Adversarial Attacks on Parts of Speech: An Empirical Study in Text-to-Image Generation}

\author{G M Shahariar, Jia Chen, Jiachen Li, Yue Dong \\
        University of California, Riverside \\ 
        \texttt{\{gshah010,jiac,jiachen.li,yue.dong\}@ucr.edu}}

\begin{document}
\maketitle
\begin{abstract}
Recent studies show that text-to-image (T2I) models are vulnerable to adversarial attacks, especially with noun perturbations in text prompts. In this study, we investigate the impact of adversarial attacks on different POS tags within text prompts on the images generated by T2I models. We create a high-quality dataset for realistic POS tag token swapping and perform gradient-based attacks to find adversarial suffixes that mislead T2I models into generating images with altered tokens. Our empirical results show that the attack success rate (ASR) varies significantly among different POS tag categories, with nouns, proper nouns, and adjectives being the easiest to attack. We explore the mechanism behind the steering effect of adversarial suffixes, finding that the number of critical tokens and content fusion vary among POS tags, while features like suffix transferability are consistent across categories. 
We have made our implementation publicly available at - 
\url{https://github.com/shahariar-shibli/Adversarial-Attack-on-POS-Tags}.
\end{abstract}

% \url{https://anonymous.4open.science/r/POS-Attack}
\section{Introduction}

Text-to-Image (T2I) generation models such as Stable Diffusion \citep{Rombach_2022_CVPR, podell2023sdxl}, DALL-E2 \citep{ramesh2022hierarchical}, Imagen \citep{saharia2022photorealistic}, ediff-i \citep{balaji2022ediff} have made steady progress in the field of image generation by bridging the semantic gap between textual descriptions and visual representations. Unlike traditional methods reliant solely on pixel manipulation, these models leverage multi-model alignments in latent spaces to interpret and synthesize complex visual content from textual prompts. Recent studies, such as \citet{tang-etal-2023-daam}, have interpreted how cross-alignment from texts to images is transformed through text-image attribution analysis, demonstrating that different POS tags are well captured by cross-modal attention during synthesis.
%The emergence of Text-to-Image Diffusion models (TDMs) has been driven by two main factors. First, the availability of extensive multi-modal text-image datasets \citep{krishna2017visual, lin2014microsoft, thomee2016yfcc100m, schuhmann2022laion} has provided a wealth of diverse data for training these models. Second, the development of guidance mechanisms \citep{nichol2021glide, dhariwal2021diffusion, ho2022classifier, song2020score} has improved the accuracy and controllability of the generated images.

\begin{figure}[t]
  \includegraphics[scale=0.36]{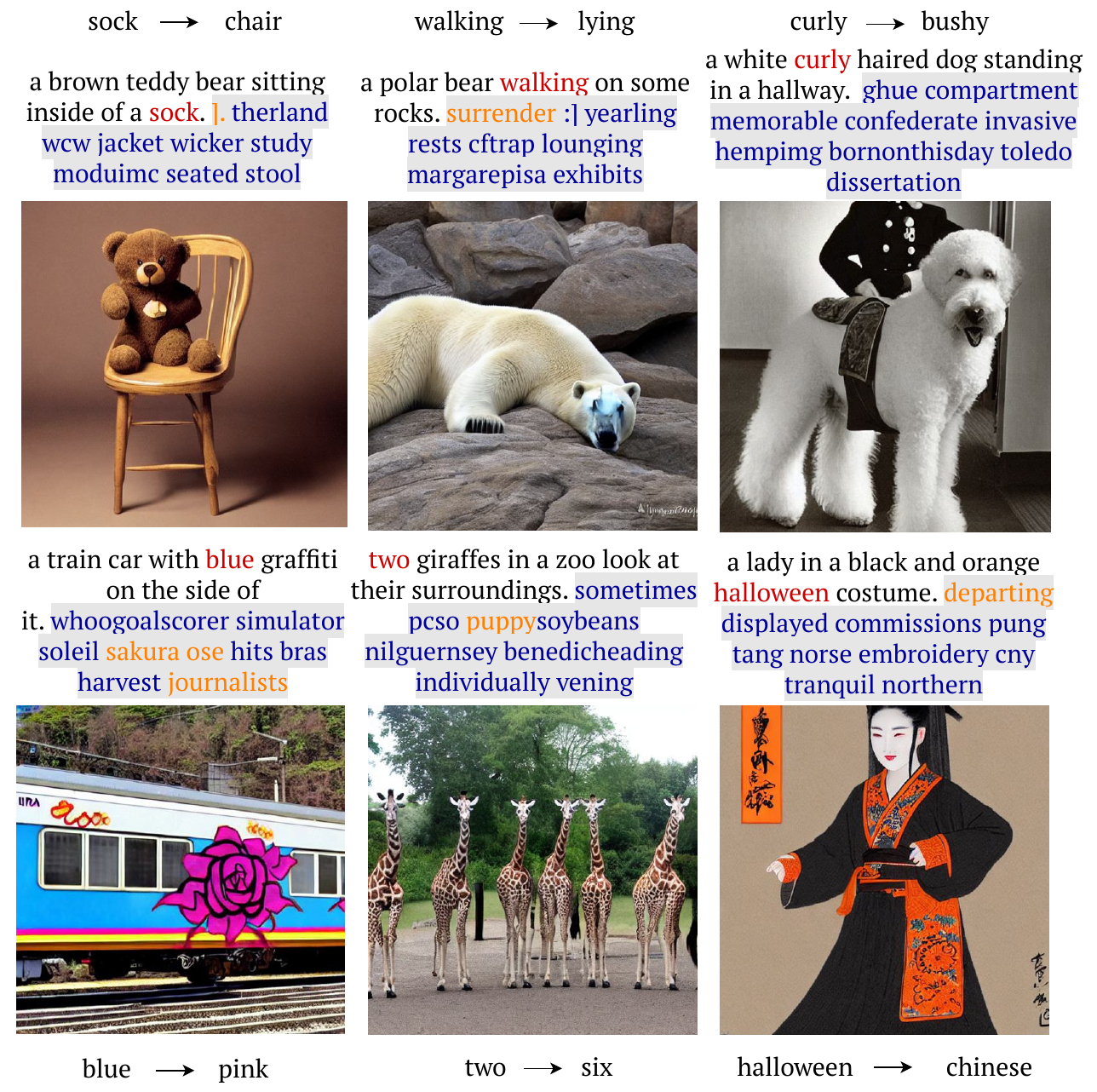}
  \caption{Examples of successful adversarial attacks on Stable Diffusion covering different POS tags drawn from our dataset. The POS tokens targeted by \colorbox{gray!10}{\textcolor{darkblue}{adversarial suffixes}} are highlighted in \textcolor{red}{red}. In addition, we observe that the attack success rate (ASR) varies significantly across POS tag categories, with features like the number of critical tokens (defined in \S \ref{sec:analysis}, non-critical tokens are highlighted in \textcolor{orange}{orange}) being highly associated with ASR.}
  \label{fig:intro-example}
\end{figure}

\begin{table}[t]
\centering
\resizebox{\columnwidth}{!}
{\begin{tabular}{c|c|c} 
\toprule
\textbf{Research Paper}       & \textbf{POS Tags to Attack}                                                                   & \textbf{Data Source}                                     \\ 
\cmidrule{1-2}\cline{3-3}
\citet{zhuang2023pilot}       & Noun                                                                                          & ChatGPT                                                  \\
\citet{liu2023discovering}    & Noun                                                                                          & ImageNet-1K                                              \\
\citet{shahgir2023asymmetric} & Noun                                                                                          & \begin{tabular}[c]{@{}c@{}}Manual\\MS-COCO\end{tabular}  \\
\citet{yang2024cheating}      & Noun                                                                                          & MS-COCO                                                  \\
\citet{yang2024sneakyprompt}  & Noun                                                                                          & ChatGPT                                                  \\
\citet{du2024stable}          & Noun                                                                                          & ImageNet-1K                                              \\ 
\cmidrule{1-2}\cmidrule{3-3}
\textbf{This work}            & \begin{tabular}[c]{@{}c@{}}Noun, Proper Noun, Adjective, \\Verb, Numeral, Adverb\end{tabular} & MS-COCO                                                  \\
\bottomrule
\end{tabular}}
\caption{Comparison of T2I adversarial attacks based on targeting parts of speech.}
\label{tab:dataset-compare}
\end{table}
On the other hand, recent research shows that T2I models are vulnerable to adversarial perturbations in text prompts, such as inserting nonsensical words \cite{milliere2022adversarial}, phrases \cite{maus2023black}, or irrelevant characters \cite{zhuang2023pilot}, which can significantly bias the generated images \citep{chefer2023attend,salman2023raising}. However, current adversarial attacks on T2I generation models, either manual heuristic-based methods \citep{zhuang2023pilot, gao2023evaluating, maus2023black} or automatic gradient-based approaches \citep{zhuang2023pilot, pmlr-v202-liang23g, liu2023discovering, shahgir2023asymmetric, yang2024cheating, yang2024sneakyprompt, du2024stable, zhai2024discovering}, are specifically targeting entities or objects (i.e., \textit{nouns}) in text prompts, neglecting other parts of speech. In this paper, we aim to answer the following two research questions:
\begin{itemize}
    \item \textbf{Q1}: Do adversarial attacks, particularly gradient-based attacks on T2I models, behave similarly when targeting different POS tag categories?
    
    \item \textbf{Q2}: Are there common or distinct features relevant to attack success rates (ASR) when targeting different POS tag categories under adversarial attacks?
\end{itemize}

To bridge the gap in analyzing attack mechanisms across different POS tag categories beyond nouns, we first created a dataset with realistic scenarios for swapping different POS tag categories with adversarial attacks. Figure \ref{fig:intro-example} provides a few examples drawn from our dataset covering six POS tags from \citet{tang-etal-2023-daam}: \textit{noun, adjective, verb, adverb, numeral}, and \textit{proper noun}, with adversarial suffixes that successfully mislead T2I models into generating images related to the targeted attribute. Creating such a dataset is non-trivial, as \citet{shahgir2023asymmetric} noted that ASR to T2I models might be affected by internal bias rather than the attack itself; we tried to minimize such biases when creating the dataset. To the best of our knowledge, there is currently no dataset available for analyzing adversarial attacks on POS tags other than nouns (refer to Table \ref{tab:dataset-compare}).

We conduct targeted adversarial attacks over POS tag categories with a gradient-based token searching algorithm specifically designed for T2I models to effectively navigate the larger vocabulary size of the T2I text encoder \citep{shahgir2023asymmetric}. The attack objective is to create an adversarial prompt that causes a target POS token to appear in the generated image while ensuring the original POS token from the input prompt does not. We observe that the ASR differs significantly across different POS tag categories. Nouns, proper nouns, and adjectives are the easiest to attack, with increasing difficulties, while the other three categories, with the same gradient-based attack, offer almost no success, whether in restricted (preventing the target token's POS tag attribute from appearing in the adversarial suffix) or unrestricted settings.

This observation led us to further investigate whether there are features associated with these differences in ASR across POS categories. Through extensive experiments, we discovered a correlation between the number of critical tokens in adversarial suffixes and the attack success rate across different POS categories. Critical tokens are those whose removal from the adversarial suffix renders the attack unsuccessful.

Additionally, the results from our ablation study reveal that adversarial suffixes, while steering the generation of target attributes, often fail to completely remove the original attribute. For example, when attempting to change a purple grape to a green one using adversarial suffixes, the resulting image often shows a mixed color. In contrast, with noun attacks, both objects can be generated, whereas with verbs, it is difficult to mix or generate both original and swapped tokens. This varying ease of content fusion across POS categories may also contribute to the differences in ASR.

Furthermore, we identified a general feature shared across different POS categories: the transferability of the attack suffix. Similar to nouns, the adversarial suffixes found are universally transferable to other input prompts with the same attributes. This means an adversarial suffix can transform multiple input prompts with different attributes into the same target attributes in the generated images. For instance, we found that the same adversarial suffix targeting a `blue' cup can steer the model to generate images of a blue cup across multiple input prompts with different original colors (e.g., red, yellow, orange).  
% \yue{rewrote abstract and intro based on our earlier discussion today. can you correct and point these examples and observation into the right part when you rewrite attack success mechanism make sure to include examples demonstrate each of these points. }

% \begin{enumerate}[align=left, labelsep=0cm]
%     \item Given that current adversarial attack datasets only focus on targeting nouns, we construct a dataset to study how existing gradient-based search approaches behave when targeting different POS categories (Section \ref{sec:dataset}).

%     \item In both unrestricted and restricted settings for targeted attacks, we observed that the attack success rate varies drastically based on the POS categories (Section \ref{sec:evaluation}).

%     \item We investigate the reasons behind the varying success rates and found that adversarial suffixes do not uniquely contribute to ASR in attacks, but there is an association between the number of critical tokens and the attack success rate (Section \ref{sec:analysis}).
% \end{enumerate}

\section{Related Work}
\textbf{Text-to-Image Diffusion Models}. \citet{nichol2021glide} formalized the initial text-to-image (T2I) diffusion model (GLIDE) that substituted class labels with text in class-conditioned diffusion models (i.e. Ablated Diffusion Models \citep{dhariwal2021diffusion}). The authors explored two types of text conditioning methods: classifier guidance and classifier-free guidance (CFG). \citet{saharia2022photorealistic} proposed Imagen by following the classifier-free guidance (CFG) of GLIDE for T2I generation. They utilized pre-trained large language models (LLMs) as the text encoder and found that scaling up language models is more efficient in improving sample fidelity and aligning images with text. \citet{ramesh2022hierarchical} created DALL-E2, a T2I generation model capable of sequentially generating images using text embeddings to guide the process. They achieved this by training a generative diffusion decoder to reverse the image encoding process of CLIP \citep{radford2021learning}. \citet{Rombach_2022_CVPR} developed the Latent Diffusion Model (LDM) by incorporating denoising methods within the latent space of pre-trained autoencoders and improving the U-Net architecture with the cross-attention mechanism. Stability AI has utilized the LDM framework to create and introduce a variety of text-to-image diffusion models called the Stable Diffusion series.

\noindent\textbf{Adversarial Attacks on T2I Models}. Existing research on adversarial attacks on T2I models primarily falls into two categories: query or heuristic-based and gradient-based. Within the first category, recent studies have explored the excessive sensitivity of T2I diffusion models to minor changes in text prompts. \citet{maus2023black} introduced a query-based attack that discovers prepended prompts capable of causing T2I diffusion models to generate specific image categories. \citet{zhuang2023pilot} targeted the text encoder of diffusion models by appending extra nonsensical characters to the input prompt using a genetic algorithm. \citet{gao2023evaluating} first identified keywords based on their impact on the generation distribution and then applied character-level substitutions, such as typos, glyphs, and phonetic variations. In the second category, there has been a recent increase in gradient-based adversarial attacks targeting the text encoder of T2I models. \citet{liu2023discovering} introduced a gradient-guided optimization process to refine a continuous token embedding, using gradients to navigate the prompt space and identify failure cases. \citet{yang2024cheating} explored a focused targeted attack that adds target objects while removing original ones, and developed MMP-Attack, which incorporates multi-modal features. \citet{du2024stable} proposed Auto-attack on Text-to-image Models (ATM), which automatically generates attack prompts that resemble clean prompts by replacing or adding words. \citet{shahgir2023asymmetric} applied gradient-based token perturbation methods to replace entities in the prompt with adversarial suffix tokens. We adopt the gradient attack proposed by \cite{shahgir2023asymmetric} because it aligns with our attack objectives and demonstrates strong performance in targeting nouns. 

\section{Dataset Creation}\label{sec:dataset}
In this section, we outline the procedure for constructing our dataset. We first specify the dataset source and then describe the steps involved in its construction.

\noindent\textbf{Data Collection}. The first obstacle we encountered in evaluating adversarial attacks across different POS categories beyond nouns was that there was no existing dataset for fair comparison. Table \ref{tab:dataset-compare} compares existing adversarial attack datasets by size, parts of speech covered, and data sources. To construct our dataset, we chose MS-COCO \cite{lin2014microsoft} as the data source for its diverse and complex captions making it suitable for testing the robustness of SD. In the train split of MS-COCO, each image has five captions. We collected only the first caption among the five resulting in 118,287 rows.

\noindent\textbf{Input Prompts Selection}. We identified the POS tags in each caption from the initially collected data using the NLTK library \cite{bird-2006-nltk} and a pre-trained POS tagging model \cite{sajjad-etal-2022-analyzing}. We only focused on six parts of speech tags: noun, verb, adverb, adjective, numeral, and proper noun. For each POS tag, we then randomly selected 20 unique captions, each containing at least one word from the corresponding POS tag, to be used as input prompts.

\noindent\textbf{Target Prompts Generation}. For each input prompt of every POS tag, we generated five target prompts, resulting in 100 prompt pairs per POS tag. Each input and target prompt differed by only one word, with the target words chosen from a pool of candidate words. The process of generating target prompts starts by extracting the POS-tagged word from the input prompt using the same NLTK library and pre-trained POS tagging model employed during the input prompt selection. Then, we compile a set of candidate words by gathering other words of the same POS category, identifying antonyms to introduce variety, \verb|[MASK]| prediction to acquire the top-5 words, and exploring the CLIP token embedding space to find the top-k distant neighbors of the word. To extract antonyms, we use the NLTK library and the WordNet database \cite{fellbaum2010wordnet}. For \verb|[MASK]| prediction, we employ BERT \cite{devlin-etal-2019-bert} as a masked language model. To identify the farthest neighbor tokens in the vocabulary space, we calculate the cosine similarity between the extracted input word embedding and the embeddings of other tokens in the vocabulary, selecting the top 100 tokens with the lowest cosine similarity scores. These candidate words are then filtered to ensure they retain the same POS while removing synonyms, subwords, and substrings to maintain relevance and avoid redundancy. Using these filtered candidate words, we then generate ten candidate prompts ranked highest through \verb|[MASK]| prediction probabilities. These prompts are subsequently ranked based on their perplexity scores, which measure the fluency and coherence of the prompts. The perplexity score is calculated using the GPT-2 model \cite{radford2019language}. Finally, the five prompts with the lowest perplexity scores, indicating the highest quality, are selected as the final target prompts. We repeat this process for each of the six POS tags, resulting in a total of 600 prompt pairs.

\noindent\textbf{Annotator Recruitment}. Our study involves two annotation tasks: dataset annotation and attack success evaluation. For these tasks, we chose two annotators with expertise and research experience in vision and language-related tasks. We chose them from a group of five candidates based on their trustworthiness scores \citep{price2020six}, which were determined through an assessment. We presented them with 30 image-text pairs and asked whether the image accurately reflected the text description (Yes/No). From our dataset, we randomly chose 20 text prompts and generated one image per prompt using SD. In addition, we created 10 text prompts with the help of ChatGPT using the prompt ``Generate 10 simple scenes for text-to-image generative model'' and then using SD generated one image per prompt. These 10 image-text pairs served as control samples, which were unknown to the participants in advance. Upon completion of the task, we assessed the number of correctly labeled control samples for each candidate. Candidates who achieved a trustworthiness score exceeding 90\% were selected as annotators.

\noindent\textbf{Dataset Annotation}. We assigned one annotator the task of assessing the meaningfulness of the generated target prompts. The annotator was provided with 600 prompt pairs. For each prompt, we also presented the annotator with 10 candidate target words generated using ChatGPT. We used the prompt ``Replace \verb|[MASK]| with the most probable 10 words in the following text: '' to generate candidate target words by ChatGPT. If the target prompt generated from our pipeline appeared meaningful and the annotator considered it visually representable, we instruct the annotator to retain it; otherwise, we ask to replace the corresponding word with an alternative from the pool of ChatGPT-generated words. Out of the 600 prompt pairs, the annotator opted to replace 97 target prompts. 

\section{Experiment}\label{sec:evaluation}
In this section, we outline the gradient-based adversarial attack method, describe the experimental setup, and report the results to assess the effectiveness of the attack.
%\subsection{Problem Formulation} We assume that we have a token vocabulary set $V = \{t_{1}, t_{2}, .... , t_{|V|}\}$. We start with an original prompt $C$ consists of $n$ tokens, $C = \{c_{1}, c_{2}, c_{3}, ..., c_{n}\}$ and $C \in V$. Let $C^\prime$ represent the adversarial prompt, which has $n+m$ tokens and also $C^\prime \in V$. The additional $m$ tokens we want to optimize will be concatenated to $C$ to form the adversarial prompt $C^\prime$. %For targeted attacks, we require the input and target prompt pairs to differ by exactly one POS token and to be visually distinct. Hence, our objective is to find suffix tokens that, when concatenated to the original prompt, direct the T2I diffusion model to generate an image containing the target POS token but is visually unrelated to the original prompt. 
\subsection{Attack Method}
Gradient-based attacks on Stable Diffusion \citep{zhuang2023pilot,shahgir2023asymmetric,yang2024cheating,yang2024sneakyprompt,du2024stable} utilize the gradient information to perturb the input prompt in a way that maximizes the divergence from the intended output, effectively manipulating the image synthesis process. While previous studies have predominantly focused on nouns, our analysis extends this approach to other parts of speech by applying the gradient-based attack framework proposed by \cite{shahgir2023asymmetric}. The process of such an attack on T2I models generally starts with an initial prompt, which is modified iteratively to create an adversarial prompt that maximizes a predefined score function. This involves embedding the target prompt and the adversarial prompt using a token embedder and processing them through a text encoder. The core mechanism focuses on creating multiple candidate prompts by replacing tokens and computing the top-k token candidates. The best candidate prompt, which maximizes the score function, is selected, and the gradient of the loss function concerning the adversarial prompt is used to iteratively refine the prompt. This iterative optimization adjusts the adversarial prompt to gradually increase the discrepancy between the model's output for the target prompt and the adversarial prompt, effectively fooling the T2I generation model. Further details of the attack are provided in Appendix \ref{app:B}. We conducted the targeted attack under two distinct settings: with and without restrictions. In the unrestricted setting, we allow the adversarial prompt to include the target token or its sub-tokens as suffix tokens. However, in the restricted attack scenario, we confine the appearance of the target token within the adversarial prompt by constraining all possible substrings of the target token.

\subsection{Experimental Setup} 
We followed the setup of \citet{shahgir2023asymmetric} and conducted the attack five times for each pair, with 100 steps per run, employing 10 adversarial tokens. For each step, we selected the top 256 tokens as candidate tokens and generated 512 new prompts by randomly substituting tokens using these candidates. Subsequently, we generated seven images per attack, resulting in the evaluation of a total of 21,000 generated images (600 pairs, 5 runs, and 7 images per run). During image generation, we set the resolution to 512 $\times$ 512, the number of inference steps to 50, and the scale of classifier-free guidance to 7.5. As the victim model, we utilized Stable Diffusion v1.5 (SD v15) for both image generation and performance assessment, leveraging a pre-trained CLIP model trained on a dataset comprising text-image pairs. All experiments (attack execution, evaluation, and image generation) were conducted using a single Nvidia RTX 3090 GPU, totaling approximately 600 GPU hours. The execution time to attack a single input-target prompt pair is approximately 8 minutes.

\subsection{Evaluation Metrics}
\textbf{Attack Success Rate}. We consider an attack successful if the image generated by the adversarial prompt matches the target text; otherwise, we consider it unsuccessful. Since we generate 7 images per adversarial prompt, to measure the attack success rate (ASR), we consider the attack as successful if at least 4 images have a higher matching score than a threshold. Following \cite{shahgir2023asymmetric}, we set this threshold value at 3.41. We determine the matching score by calculating the difference between the CLIP score of the input prompt and the generated image, and the CLIP score of the target text and the generated image.
% \yue{how do we determine matching? do we have a threashold if we are clip score?} \shah{added}. 
CLIP score measures the cosine similarity between the visual CLIP embedding of an image and the textual CLIP embedding of a text. For each input-target prompt pair, we run the attack five times, generating five adversarial prompts, and consider the attack successful if at least one of them succeeds.

\noindent\textbf{Semantic Shift Rate}. For a quantitative measure to evaluate the efficacy of adversarial suffix tokens, we utilized SemSR (Semantic Shift Rate) \citep{zhai2024discovering} which measures the semantics between a generated image and a text prompt. SemSR utilizes CLIP's multi-modal embedding space and computes the similarity in semantics between a generated image and a prompt using cosine similarity. This metric quantifies the displacement in the vector space of the generated image after appending adversarial suffix tokens compared to the image generated using the input prompt. Since the amount of deviation necessary to attain diverse target semantics differs, it is adjusted by the maximum deviation. The SemSR equation is provided below:
\begin{equation}
    SemSR = \frac{CS(E_{I_{a}}, E_{P_{a}}) - CS(E_{I_{i}}, E_{P_{i}})}{CS(E_{I_{t}}, E_{P_{t}}) - CS(E_{I_{i}}, E_{P_{i}})}
\end{equation}
where $CS$ denotes $CLIP\_Score$, $I_{a}$ represents the generated image from the adversarial prompt $P_{a}$, $I_{i}$ denotes the generated image from the input prompt $P_{i}$, and $I_{t}$ denotes the generated image from the target prompt $P_{t}$. For a single input-target prompt pair, we measure the average of SemSR scores over five runs.

\subsection{Results}
In Figure \ref{fig:attack-example}, we showcase a few examples of images generated through both the unrestricted and restricted attack methods. Table \ref{tab:asr-semsr} displays the average attack success rate (ASR) and average semantic shift rate (SemSR) over all the prompt pairs for each POS tag under both attack conditions. Below, we present both quantitative analysis and human evaluation of our experiments.

\begin{table}[hb]
\centering
\resizebox{\columnwidth}{!}{
\begin{tabular}{l|cc|cc} 
\toprule
\multirow{2}{*}{\textbf{POS Tag}}                    & \multicolumn{2}{c|}{\textbf{Unrestricted Attack}} & \multicolumn{2}{c}{\textbf{Restricted Attack}}  \\ 
\cmidrule{2-5}
                                                     & \textbf{ASR}  & \textbf{SemSR}                    & \textbf{ASR}  & \textbf{SemSR}                  \\ 
\midrule
Noun                                                 & \textbf{0.65} & 1.4394                            & \textbf{0.51} & 1.3884                          \\
Proper Noun                                            & 0.40          & 0.8955                   & 0.31          & 0.8606                          \\
Adjective                                                 & 0.29          & \textbf{2.0929}                            & 0.24          & 1.1181                 \\
Verb                                               & 0.15          & 1.5963                            & 0.12          & \textbf{1.9121}                          \\
Numeral                                              & 0.13          & 1.9246                            & 0.11          & 1.5943                          \\
Adverb & 0.03          & 0.9313                            & 0.01          & 1.0077                          \\
\bottomrule
\end{tabular}}
\caption{Average Attack Success Rate (ASR) and average Semantic Shift Rate (SemSR) of both unrestricted and restricted attack on each POS Tag. The higher the values, the better. The highest values are \textbf{bold} marked.} 
% \yue{ASR and semSR seem to be inconsistent with each other, which one shall we trust?} \shah{added at a later paragraph}}
\label{tab:asr-semsr}
\end{table}

\noindent\textbf{Quantitative Evaluation}. Table \ref{tab:asr-semsr} presents the ASR and SemSR metrics, which are the average values across 100 data points for each POS tag. Higher ASR and SemSR values indicate better performance. From the table, we observe that in the case of the unrestricted attack, both ASR and SemSR surpass those of the restricted attack except for verb and adverb POS tags. This suggests that allowing the target token to be part of the concatenated adversarial suffix tokens leads to greater success in adversarial attacks. Clearly, the unrestricted attack excels in producing images containing the target POS token instead of the input POS token. We also observe that in both the restricted and unrestricted attacks, nouns demonstrate higher ASR values compared to other POS tags, implying their greater vulnerability to adversarial attacks. Proper nouns and adjectives show moderate success rates, while verbs and numerals exhibit lower success rates. Adverbs, on the other hand, have the lowest success rates in both types of attacks, indicating their higher resistance to adversarial manipulation. SemSR values quantify the semantic disparity between a text and its corresponding generated image caused by an adversarial attack. Higher SemSR values signify substantial semantic shifts. By analyzing SemSR values, we find that attacking nouns and adjectives is comparatively simpler, whereas adverbs present greater difficulty. This suggests that nouns and adjectives undergo more significant semantic alterations, while adverbs experience the least. Moreover, SemSR values remain relatively stable across various POS tags for both unrestricted and restricted attacks. However, with unrestricted attack, adjectives exhibit the greatest semantic shifts, which is not the case with restricted attack. Numerals consistently show the second-highest semantic changes across both attack types.
\begin{table}
\centering
\resizebox{\columnwidth}{!}{
\begin{tabular}{c|cc|cc} 
\toprule
\multirow{2}{*}{\textbf{POS Tag}}                    & \multicolumn{2}{c|}{\textbf{Unrestricted Attack}} & \multicolumn{2}{c}{\textbf{Restricted Attack}}  \\ 
\cmidrule{2-5}
                                                     & \textbf{Input} & \textbf{Target}                  & \textbf{Input} & \textbf{Target}                \\ 
\midrule
Noun                                                 & 0.13           & 0.87                             & 0.27           & 0.73                           \\
Proper Noun                                            & 0.47           & 0.53                             & 0.53           & 0.47  
                    \\
Adjective                                            & 0.67           & 0.33                             & 0.53           & 0.47                           \\
Verb                                                 & 0.73           & 0.27                             & 0.67           & 0.20                          \\
Numeral                                              & 0.20           & 0.13                             & 0.20           & 0.13                           \\
Adverb                                               & 0.93           & 0.07                             & 0.87           & 0                           \\
\bottomrule
\end{tabular}}
\caption{Human evaluation results on the matching of input and target text with the generated images for each POS tag in both unrestricted and restricted settings.}
\label{tab:human-eval}
\end{table}

\noindent\textbf{Human Evaluation}. We evaluate the attack's effectiveness with the assistance of two annotators. We randomly choose 15 prompt pairs for each POS tag, amounting to a total of 90 prompt pairs for both unrestricted and restricted attack settings. Each prompt pair is presented with 7 images to the annotators (as 7 images were generated per run in our experiments), who then assess whether at least 4 images closely align with either the target prompt or the input prompt (Yes/No). We collect evaluations from the annotators using a Google Form (Appendix \ref{app:K}), which includes the generated image and two checkboxes for the input text and target text. We determine the score of an annotator by the number of prompt pairs they classify as a match. Since there are two evaluators, we calculate the average of their scores and present the results in Table \ref{tab:human-eval}. The table indicates that annotators agree that verbs, adverbs, and numerals are more resistant to adversarial attacks. In the case of numerals, the annotators reported that the majority of the post-attack generated images do not align with either the target or the input prompts. We observe that unrestricted attack tends to generate images that more closely match the target prompt than the restricted attack. We used Cohen's Kappa ($\kappa$) metrics \cite{cohen1960coefficient} to measure annotator agreement on target text-image matching, obtaining scores of 0.796 for unrestricted and 0.745 for restricted settings, which indicate a high degree of agreement.

From Table \ref{tab:asr-semsr} and \ref{tab:human-eval}, we observe that the average ASR shows a strong positive correlation with human evaluation of target text-image matching in both the unrestricted setting (Pearson = 0.988 and Spearman = 1.00) and the restricted setting (Pearson = 0.980 and Spearman = 0.986). On the other hand, the average SemSR exhibits a very weak negative correlation with human evaluation in both unrestricted attack scenario (Pearson = -0.126 and Spearman = -0.143), and restricted attack scenario (Pearson = -0.176 and Spearman = -0.087). Given that the average ASR has higher correlations with human judgment in both settings, it is more reliable than average SemSR for evaluating the success of attacking POS tags. Therefore, we use ASR to evaluate attack success in all subsequent sections.
% \yue{maybe the human evaluation in table 3 is giving the answer on whether we should use ASR or SemSR, can we run some correlation stats on these two, and then say one automatic measure is more reliable than the other for POS attack success evaluation? (it seems that you are using ASR for the rest of the sections, so probably ASR should have higher correlations with human judgetment.} \shah{added}

% \begin{figure}[t]
%   \centering
%   \includegraphics[scale=0.45]{Figures/suffix_separated.pdf}
%   \caption{Token level visual depiction of an adversarial suffix in restricted setting. $\oplus$ indicates concatenation operation.}
%   \vspace{-1.5em}
%   \label{fig:token_separated}
% \end{figure}

%% critical and non-critical table
\begin{table*}
\centering
\resizebox{\textwidth}{!}{
\begin{tabular}{c|ccc|ccc} 
\toprule
\multirow{2}{*}{\begin{tabular}[c]{@{}c@{}}\textbf{}\\\textbf{POS}\\\textbf{Tag}\end{tabular}} & \multicolumn{3}{c|}{\textbf{Unrestricted }}                                                                                                                                                                                                                                                                       & \multicolumn{3}{c}{\textbf{Restricted }}                                                                                                                                                                                                                                                                           \\ 
\cmidrule{2-7}
                                                                                               & \begin{tabular}[c]{@{}c@{}}\textbf{Number of }\\\textbf{Successful }\\\textbf{Attack}\end{tabular} & \begin{tabular}[c]{@{}c@{}}\textbf{Avg no. of}\\\textbf{critical }\\\textbf{tokens}\end{tabular} & \begin{tabular}[c]{@{}c@{}}\textbf{Avg ASR }\\\textbf{by removing}\\\textbf{critical tokens}\end{tabular} & \begin{tabular}[c]{@{}c@{}}\textbf{Number of }\\\textbf{Successful }\\\textbf{Attack}\end{tabular} & \begin{tabular}[c]{@{}c@{}}\textbf{Avg no. of}\\\textbf{critical }\\\textbf{tokens}\end{tabular} & \begin{tabular}[c]{@{}c@{}}\textbf{Avg ASR }\\\textbf{by removing}\\\textbf{critical tokens}\end{tabular}  \\ 
\midrule
Noun                                                                                           & 65                                                                                                 & 7.800                                                                                            & 0.195                                                                                                     & 51                                                                                                 & 8.902                                                                                            & 0.136                                                                                                      \\
Proper Noun                                                                                    & 40                                                                                                 & 8.175                                                                                            & 0.175                                                                                                     & 31                                                                                                 & 8.935                                                                                            & 0.115                                                                                                      \\
Adjective                                                                                      & 29                                                                                                 & 7.862                                                                                            & 0.173                                                                                                     & 24                                                                                                 & 8.960                                                                                            & 0.111                                                                                                      \\
Verb                                                                                           & 15                                                                                                 & 8.200                                                                                            & 0.166                                                                                                     & 12                                                                                                 & 9.000                                                                                            & 0.076                                                                                                      \\
Numeral                                                                                        & 13                                                                                                 & 8.615                                                                                            & 0.150                                                                                                     & 11                                                                                                 & 9.180                                                                                            & 0.034                                                                                                      \\
Adverb                                                                                         & 3                                                                                                  & 9.000                                                                                            & 0.078                                                                                                     & 1                                                                                                  & 10.000                                                                                           & 0                                                                                                          \\
\bottomrule
\end{tabular}}
\caption{Comparison of the average attack success rates (ASR) by removing critical tokens across different POS tags, under all unrestricted and restricted successful attack examples.}
\label{tab:critical-and-non-critical-analysis}
\end{table*}

\section{Attack Success Mechanism}\label{sec:analysis}
In this section, we explore the mechanism behind the steering effect of adversarial suffixes. We identify (a) features that vary across POS categories and explain differences in attack success rates (ASR), such as the number of critical tokens and content fusion, and (b) features that are consistent across different POS categories and do not explain variations in ASR rates, but provide general insights such as suffix transferability.

\noindent\textbf{Correlation between the number of critical tokens in adversarial suffixes and ASR}. A successful attack demonstrates that appending an adversarial suffix to an input prompt effectively shifts the text embedding toward the target prompt, highlighting the significant role of the suffix tokens. To investigate, we tokenized several adversarial suffixes, generated an image for each token to isolate their contributions, and found that some tokens generate images associated with the target POS token. This observation led us to identify the most contributing tokens within adversarial suffixes. We define ``critical tokens'' as those whose removal causes the attack to fail.
\begin{figure}[h]
  \centering
  \includegraphics[scale=0.29]{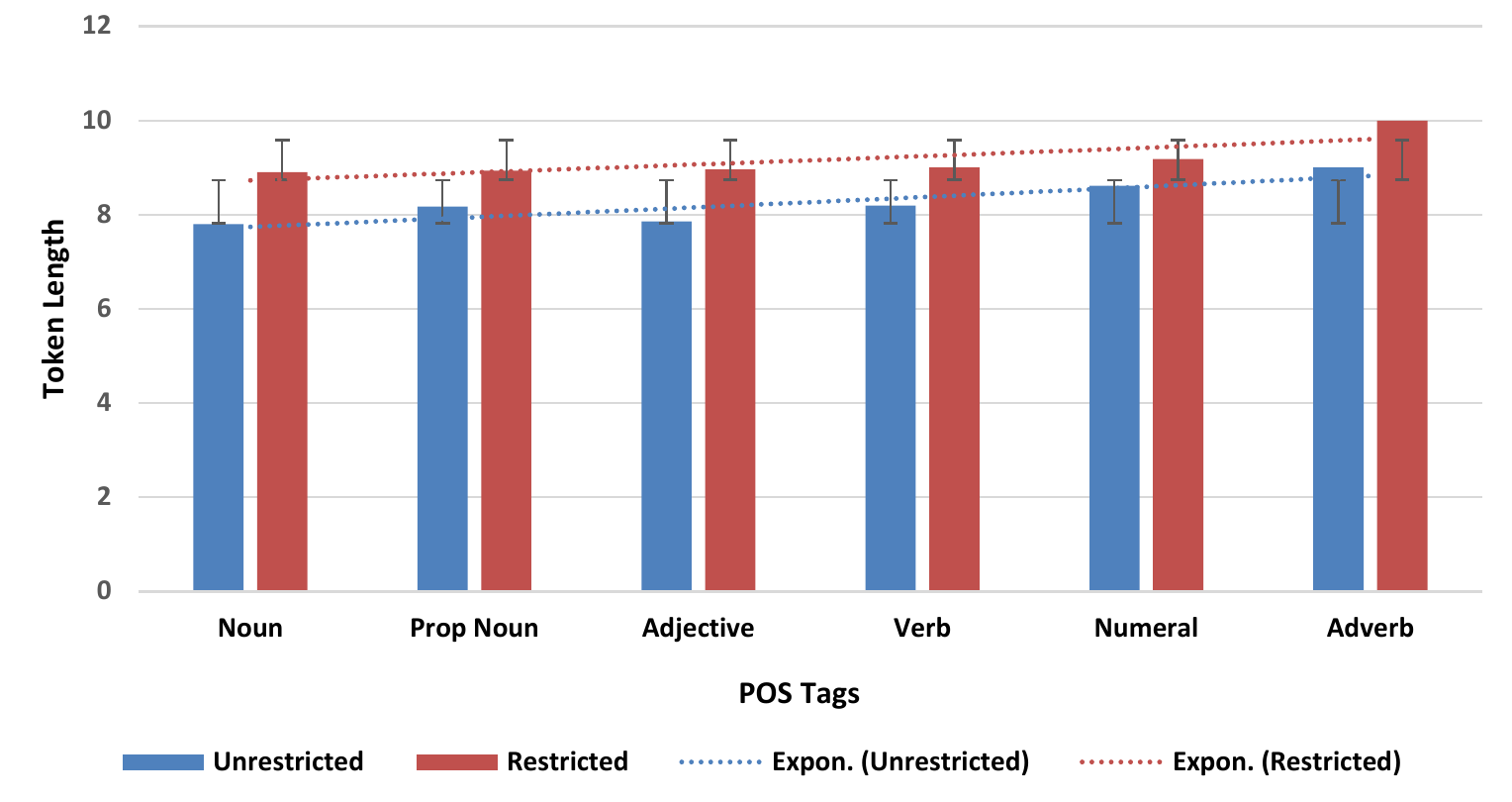}
  \caption{Average length of critical tokens across different POS tags in unrestricted and restricted settings. Exponential trend lines are included for both settings to highlight the general pattern.}
  \label{fig:critical_token_bar}
\end{figure}
To determine critical tokens in a suffix, we generated all possible combinations of replacing suffix tokens with <|endoftext|> token. 
For each combination, we generated an image and queried the pre-trained vision-language model BLIP\footnote{\url{https://huggingface.co/Salesforce/blip-vqa-capfilt-large}} \cite{li2022blip} to check if the generated image matched the target prompt. We identified the combination with the fewest tokens replaced by <|endoftext|> and considered those tokens as critical since their absence leads to an unsuccessful attack. Tokens not replaced by <|endoftext|> were considered non-critical. 

We present the average number of critical tokens across all POS tags in Table \ref{tab:critical-and-non-critical-analysis} and compare the lengths of critical tokens in Figure \ref{fig:critical_token_bar}. The number of critical tokens is generally higher across POS categories in both attack settings.  However, the restricted setting shows significantly higher numbers of critical tokens, as the absence of the target word necessitates other tokens to compensate and maintain the attack's effectiveness.
We find that adverbs, numerals, and verbs are more resistant to adversarial attacks due to their dependency on the high number of critical tokens in the suffixes. This prompted us to explore whether every critical token within a suffix contributes equally to the attack's success. Therefore, removing some or all critical tokens from the suffixes should notably decrease the ASR. To test this in both settings, we removed critical tokens from the suffixes in all possible combinations while keeping the non-critical tokens unchanged. We then calculated the ASR for each combination by querying BLIP and took the average. We find that the ASR significantly decreases across POS categories. Table \ref{tab:critical-and-non-critical-analysis} shows that adverbs, numerals, and verbs are the hardest to attack due to their reliance on a higher number of critical tokens, resulting in a significantly lower ASR when these tokens are removed. However, nouns, proper nouns, and adjectives are relatively easier to attack. Despite having a substantial number of critical tokens, the ASR for these categories remains moderately high when critical tokens are removed but still shows a significant drop. For instance, in the unrestricted setting, the attack success rate drops from 65 (total successful attacks) to around 13 (0.195 * 65) when critical tokens are removed. Thus, we conclude that the number of critical tokens in adversarial suffixes is highly associated with ASR. Some POS tags are harder to attack because the attack algorithm must find adversarial suffixes with a higher number of critical tokens.

\noindent\textbf{Ease of content fusion}. We observe that while adversarial suffixes steer the generation of target attributes, they often fail to completely remove the original token. This results in images generated by the Stable Diffusion containing both the input and target attributes, a phenomenon we refer to as content fusion. We found that content fusion across different POS categories decreases with decreasing ASR. For example, when attempting to change a noun like ``car'' to ``motorcycle'', the resulting image often contains both the car and the motorcycle. For a proper noun, changing a ``Santa costume'' to a ``Halloween costume'' might result in a Santa costume with Halloween-themed colors. With adjectives, trying to change a ``white swan'' to a ``black swan'' can lead to an image of a swan that is both black and white. We showcase examples of adjective fusion in Appendix \ref{app:G}. In contrast, verbs are harder to mix or generate together; for instance, it is difficult to create an image where a person is both standing and lying down. Similarly, in the case of numerals, attempting to change ``three apples'' to ``five apples'' often fails to produce an image with both three and five apples. With adverbs, changing ``running quickly'' to ``running slowly'' does not result in an image that simultaneously depicts both quick and slow running. We posit that this varying ease of content fusion is due to the number of critical tokens associated with ASR. In categories like nouns, proper nouns, and adjectives, where the number of critical tokens is relatively lower, fusion is easier. However, in categories with a higher number of critical tokens, such as verbs, numerals, and adverbs, fusion is not possible.

\noindent\textbf{Suffix Transferability}. 
We discovered a common feature across different POS categories: the transferability of adversarial suffixes. We observed that the identified adversarial suffixes can universally transfer to other input prompts within the same POS tag. This indicates that a single adversarial suffix can convert various input prompts with distinct attributes into images with the same target attributes. For example, an adversarial suffix targeting the noun ``motorcycle'' can prompt the model to generate motorcycle images from diverse noun prompts like ``plane'', ``car'', and ``bird''. We present some examples in Appendix \ref{app:E} (Figure \ref{fig:suffix-transfer}). Additionally, to explain why such universal transferability works,  we demonstrate by following the approach of \cite{du2024stable} that the adversarial suffix alone can dictate the output of the Stable Diffusion by steering the generated image toward the target prompt. At first, we divide a successful adversarial prompt into two segments: the input prompt and the adversarial suffix. Then we extract text embeddings of both the input prompt and the suffix separately. This step ensures that each segment is processed into its own embedding without influence from the other segment. The embeddings of the input prompt and the suffix are then concatenated. Concatenation ($\oplus$) means joining these two embeddings into a single combined embedding that the Stable Diffusion will use for image generation. We observe that the final image generated by Stable Diffusion using the concatenated text embedding matches the target prompt. We repeat this procedure for all successful attack examples and find consistent results across all POS tags. Further details can be found in Appendix \ref{app:E}.

\section{Conclusion}
In this study, we evaluate a gradient-based adversarial attack aimed at six POS tags within text prompts in both unrestricted and restricted attack strategies. We assess the impact of these attacks on the Stable Diffusion, revealing valuable insights into the factors contributing to their success. Our findings reveal that nouns, proper nouns, and adjectives are particularly vulnerable to perturbation, resulting in adversarial image generation. However, we see that verbs, adverbs, and numerals exhibit a higher level of resilience against adversarial attack, exerting minimal influence on the visual output generated by the Stable Diffusion. We hypothesize that the number of critical tokens in an adversarial suffix and the ease of content fusion are primarily responsible for such resilience against attacks. We believe these findings will be valuable for enhancing the robustness of T2I generation systems.

\section{Limitations}
We utilized the Stable Diffusion model for the gradient-based attack. It is important to note that the attack approach might not generalize effectively to other closed-source T2I generation models like Imagen \cite{saharia2022photorealistic} or DALL-E2 \cite{ramesh2022hierarchical}, owing to differences in architecture, text encoder, and training data. Moreover, the metrics utilized in this study to assess the attack may not fully capture the visual plausibility or semantic accuracy of images after the attack. We evaluated the attack only on six specific POS tags, which may not encompass all possible scenarios, such as \textit{prepositions, conjunctions, interjections, articles,} and \textit{determiners}. Furthermore, the approach relies on appending suffix tokens to the original prompt, which may not always be the most optimal method for manipulating the image generation process, considering the T2I model's sensitivity to the order of tokens. As the appended adversarial suffix tokens may lack meaning, the resulting adversarial prompt found by the attack methods exhibits reduced naturalness.

\bibliography{custom}

% \clearpage
\appendix

\section*{Appendix}

\section{Preliminaries of Stable Diffusion}
Stable diffusion is a latent diffusion model comprising three key components: a Variational Autoencoder (VAE), a UNet, and a CLIP text encoder \cite{radford2021learning} for conditioning. The VAE consists of an encoder $E$ and a decoder $D$, where the encoder compresses an image $y$ into a lower-dimensional latent space representation $E(y)$, while the decoder reconstructs the image from the latent space $\bar{y} = D(E(y))$. During the T2I generation process, at first CLIP tokenizer tokenizes a text prompt into a sequence of tokens $W = \{w_{1}, w_{2}, .... {w_{n}}\}$, ensuring uniform length by padding or truncating sequences to 77 tokens for computational ease. Each token is then converted into a text representation $W_{emb}$ using the text encoder of CLIP. CLIP comprises both an image encoder and a text encoder, each responsible for encoding an image and its corresponding text description into representations that closely align with one another. 
% \begin{figure}[h!]
%   \includegraphics[scale=0.35]{Figures/fifty-steps-generation.pdf}
%   \caption{An example of step-by-step image generation using Stable Diffusion v1-5.}
%   \label{fig:fifty-steps}
% \end{figure}
As a result, the text representation $W_{emb}$ generated by CLIP's text encoder for a given text prompt is expected to contain relevant information about the images described in the prompt. Next, a random latent image representation $I_{0}$, drawn from a Gaussian distribution is created, and noise is gradually eliminated to get a noise-free representation $E(y) = I_{emb}$ through a reverse diffusion process. Guided by the latent text embedding $W_{emb}$, a UNet neural network $U(I_{0}, W_{emb}, t)$ employs a cross-attention mechanism to predict and eliminate noise from the latent space $E(y)$ at each time step $t$. The level of noise reduction is regulated by a scheduler, progressively refining image quality. Finally, the VAE decoder $D$ upscales the latent image $E(y)$ back into pixel space, resulting in a high-resolution image $\bar{y}$. 
% In Figure \ref{fig:pipeline}, we present a detailed process of generating the adversarial prompt \textit{``a white swan on a lake. black sements gaga tiazzle shares zalbraving scratches water''} using the SD model over 50 inference steps.

\begin{figure}[t]
  \centering
  \includegraphics[scale=0.6]{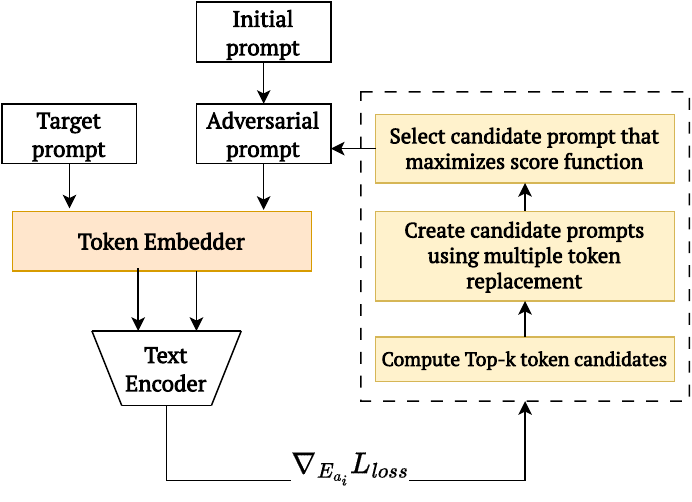}
  \caption{Schematic view of the POS-Attack pipeline. At first, hidden state representations from the CLIP text encoder using input and target token embeddings are extracted. Then, we compute loss, take gradients, and select the top-k candidate tokens for substitution. Next, we create several candidate prompts by randomly replacing multiple tokens from the pool. The candidate prompt maximizing a score function is chosen for the next optimization step.}
  \label{fig:pipeline}
\end{figure}

\begin{figure}[t]
  \centering
  \includegraphics[scale=0.41]{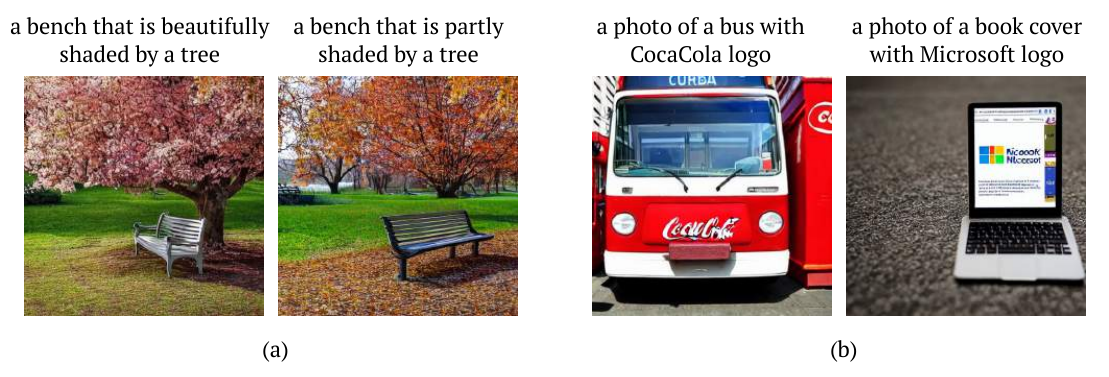}
  \caption{Examples of vulnerabilities revealed by SD model with prompts containing adverbs and proper nouns.}
  \label{fig:vulnerability_exp}
\end{figure}

\begin{figure*}
  \centering
  \includegraphics[width=\textwidth]{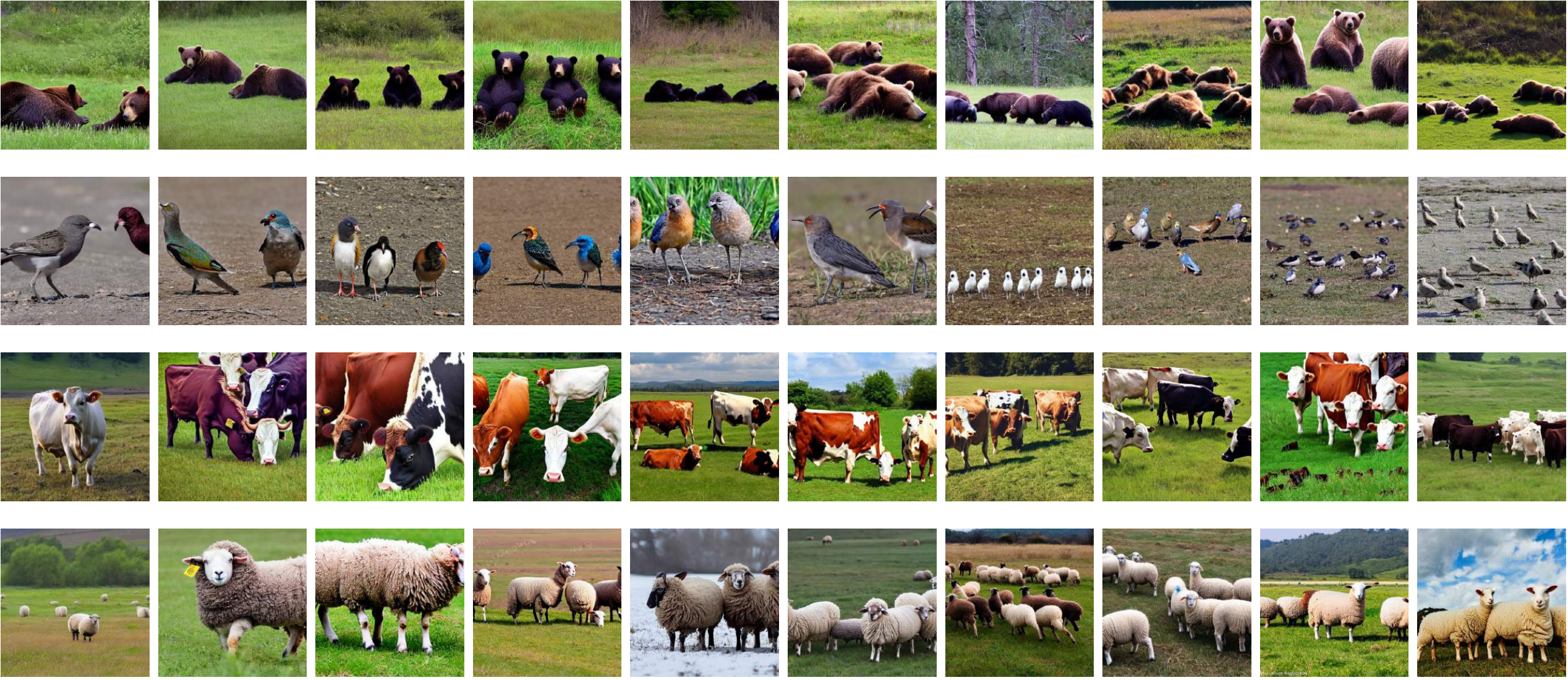}
  \caption{Examples of vulnerabilities revealed by SD model with numerals. Each row contains 10 images with numerals one to ten from left to right sequentially. The first row contains ten images of the prompt \textit{``\_\_\_ bears lying in the field''} where \_\_\_ is replaced by ``one'' to ``ten'' serially. Similarly, the second, third, and fourth rows contain prompts \textit{``\_\_\_ birds looking around while on the ground'', ''\_\_\_ cows eating grass''} and \textit{``\_\_\_ sheep roaming in the field''}.}
  \label{fig:numerical_exp}
\end{figure*}

\section{Details of Adversarial Attack}\label{app:B}
\noindent\textbf{Adversarial Prompt Generation}. We start the process by considering the input prompt as the adversarial prompt. Subsequently, we extract the embeddings for both the adversarial prompt tokens and the target prompt tokens. These embeddings are then fed into the CLIP text encoder to obtain the final hidden state representations. Following this, we compute the loss using a loss function and calculate gradients with respect to one hot token vector to determine the top-k candidate tokens for substitution. Then, we generate several candidate prompts by randomly replacing multiple tokens of the initial adversarial prompt from the pool of candidate tokens. The candidate prompt that maximizes the score function is chosen as the adversarial prompt for the subsequent optimization step. This iterative process continues for a set number of iterations until a final adversarial prompt is obtained. Notably, this attack method relies only on the text encoder and does not necessitate access to the image generation model. An illustration of the adversarial attack is depicted in Figure \ref{fig:pipeline}. From the adversary's viewpoint, the concatenated suffix tokens in the adversarial prompt should be nonsensical to humans yet encode specific semantics predetermined by the adversary.

\noindent\textbf{Loss Function}. The attack focuses on manipulating the CLIP embedding space to optimize a score function, which quantifies how much the adversarial token embeddings at an intermediate optimization stage deviate towards the target token embeddings using cosine similarity. The objective is to steer away from the embeddings of input tokens and progressively approach those of the target tokens by discovering more effective adversarial tokens. This process of maximizing the score function is similar to \citet{shahgir2023asymmetric}. To compute the loss, we adopt the negated score function. Maximizing the score is equivalent to minimizing the loss.
% \begin{equation}\label{eq1}
%     L_{loss}(E_{a}) = - S(E_{a})
% \end{equation}
% \begin{equation}\label{eq2}
%     S(E_{a}) =  \cos(E_{i}, E_{a}) - \cos(E_{t}, E{a})
% \end{equation}
% Here, $E_{i}, E_{t}$ and $E_{a}$ denote the last hidden state representations of the input prompt, target prompt, and adversarial prompt respectively.

\noindent\textbf{Gradient-based Search}. The attack employs an effective greedy coordinated gradient-based search algorithm \cite{zou2023universal}, utilizing the loss function discussed above. At each optimization step, the algorithm selects $k$ tokens with the highest negative loss and computes gradients with respect to the one-hot token vectors to identify a promising set of candidates for replacing adversarial suffix token positions.
% We can estimate the approximation of replacing the $i-th$ token $x_{i}$ by evaluating the gradient:
% \begin{equation}
%     \nabla e_{x_{i}} L_{loss} \in R^{|V|}
% \end{equation}
% Here, $R^{|V|}$ indicates that the gradient vector belongs to a real-valued vector space of size equal to the vocabulary set $V$, and $e_{x_{i}}$ indicates the one-hot vector representing the current value of the i-th token. 
New candidate prompts are generated by randomly replacing multiple token positions using the pool of token candidates, repeating this process $T$ times. Following the approach of \citet{shahgir2023asymmetric}, we initially replace all tokens and then gradually reduce the replacement rate to 20\%.
% We showcase examples of some adversarial suffixes generated using this approach in Figure \ref{fig:attack-example}.

\begin{figure*}[t]
  \centering
  \includegraphics[scale=0.65]{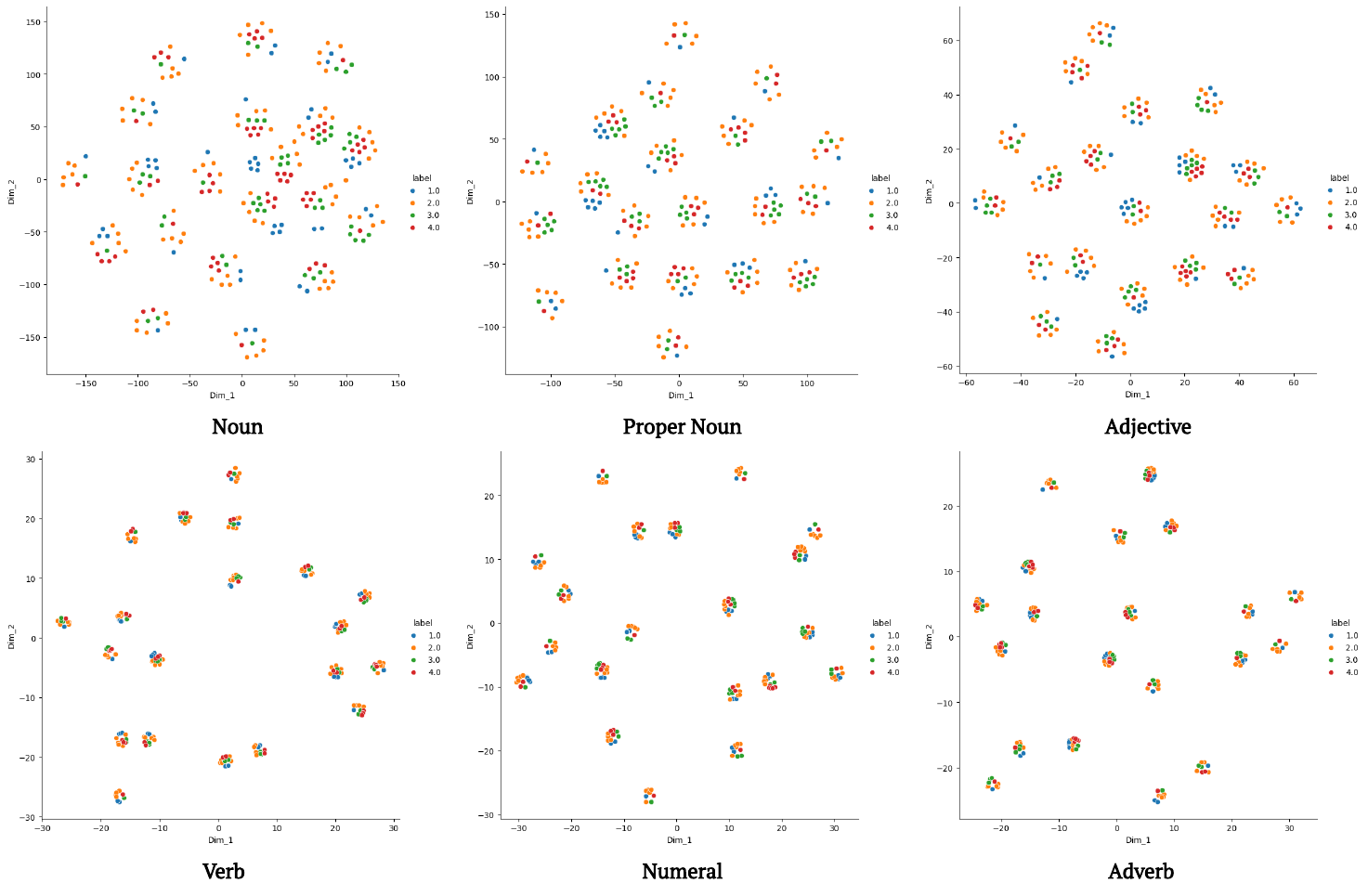}
  \caption{Visual representation of text embeddings of all input-target prompt pairs across six POS tags in 2D space. \textcolor{blue}{Blue} marker is for input prompt, \textcolor{yellow}{yellow} indicates target prompt, \textcolor{green}{green} for inherent bias inducing prompt, and \textcolor{red}{red} for prompt where the target word is removed.}
  \label{fig:embedding_plot}
\end{figure*}
\section{Vulnerabilities Observed across POS Tags}
In this section, we present some vulnerabilities observed on Stable Diffusion across a few POS tags. We noticed that the SD model inherently faces difficulty generating images from prompts that include numerals. Specifically, the model struggles to produce images with a precise count of identical objects. For instance, if the prompt is to generate an image of five birds, the SD model will fail to create exactly five birds and instead produce images with a random number of birds, such as three, four, or more than five. Examples of this issue are shown in Figure \ref{fig:numerical_exp}. Images generated by the model using prompts where the adverb tokens have shared linguistic structures, close semantic representation in the feature space, and unrelated to emotions generally have minimal impact on visual output. We show such an example in Figure \ref{fig:vulnerability_exp}(a). In this example, substituting \textit{``beautifully''} with \textit{``partly''} in the prompt \textit{``a bench that is beautifully shaded by a tree''} results in close perplexity\footnote{We utilized GPT-2 \cite{radford2019language} for perplexity score calculation.} scores for the first (128.18) and second prompts (123.49). exhibit little difference. Furthermore, we observed that the SD model struggles to generate images involving logos, such as those of Microsoft, Disney, or Google. As shown in Figure \ref{fig:vulnerability_exp}(b), instead of producing accurate images, the model generates images with misspelled words as logos.

\section{Impact of Semantic Distance on Attack Success}
In this section, we examine why certain POS tags are easier to attack by considering the impact of semantic distance. To explore this, we plotted text embeddings of all the data across six POS tags from the dataset in Figure \ref{fig:embedding_plot} using t-Distributed Stochastic Neighbor Embedding (t-SNE) \cite{van2008visualizing}. Nouns, proper nouns, and adjectives show clear clustering with visible distances between markers, indicating considerable differences in their text embeddings. This distance allows the gradient algorithm to minimize the gap from input to target, making attacks on these POS tags easier. However, for numerals, verbs, and adverbs, the markers are very close or even overlapping, indicating that the input and target prompts have similar semantic representations. This proximity hampers the algorithm's ability to optimize the distance gap, leading to lower attack success rates.  

\begin{figure}[hb]
  \includegraphics[scale=0.38]{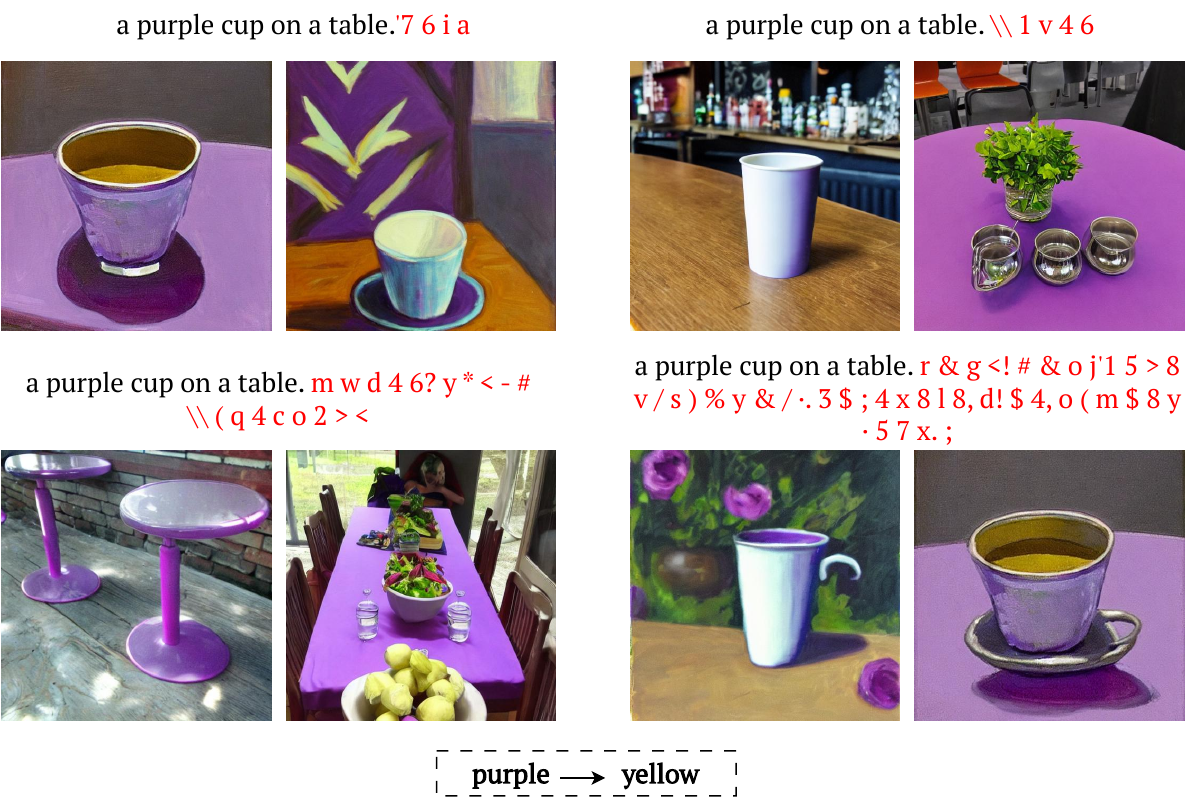}
  \caption{Examples of adversarial attack on Adjective using ASCII/non-alphabetic characters.}
  \label{fig:character-attack}
\end{figure}

\section{Attack using ASCII/non-alphabetic characters}
We evaluated the adversarial attack by limiting it to adding only ASCII or non-alphabetic characters at the end of the input prompt. However, these attempts were unsuccessful. We conducted experiments by adding 5, 20, and 50 adversarial characters to a text prompt containing an adjective token, yet there was no alteration in the resulting image. In all three scenarios, we noticed that even with the addition of characters, the SD continued to generate images identical to those generated from the original input prompt. We present a few examples in Figure \ref{fig:character-attack}.

\begin{figure}[t]
  \centering
  \includegraphics[scale=0.42]{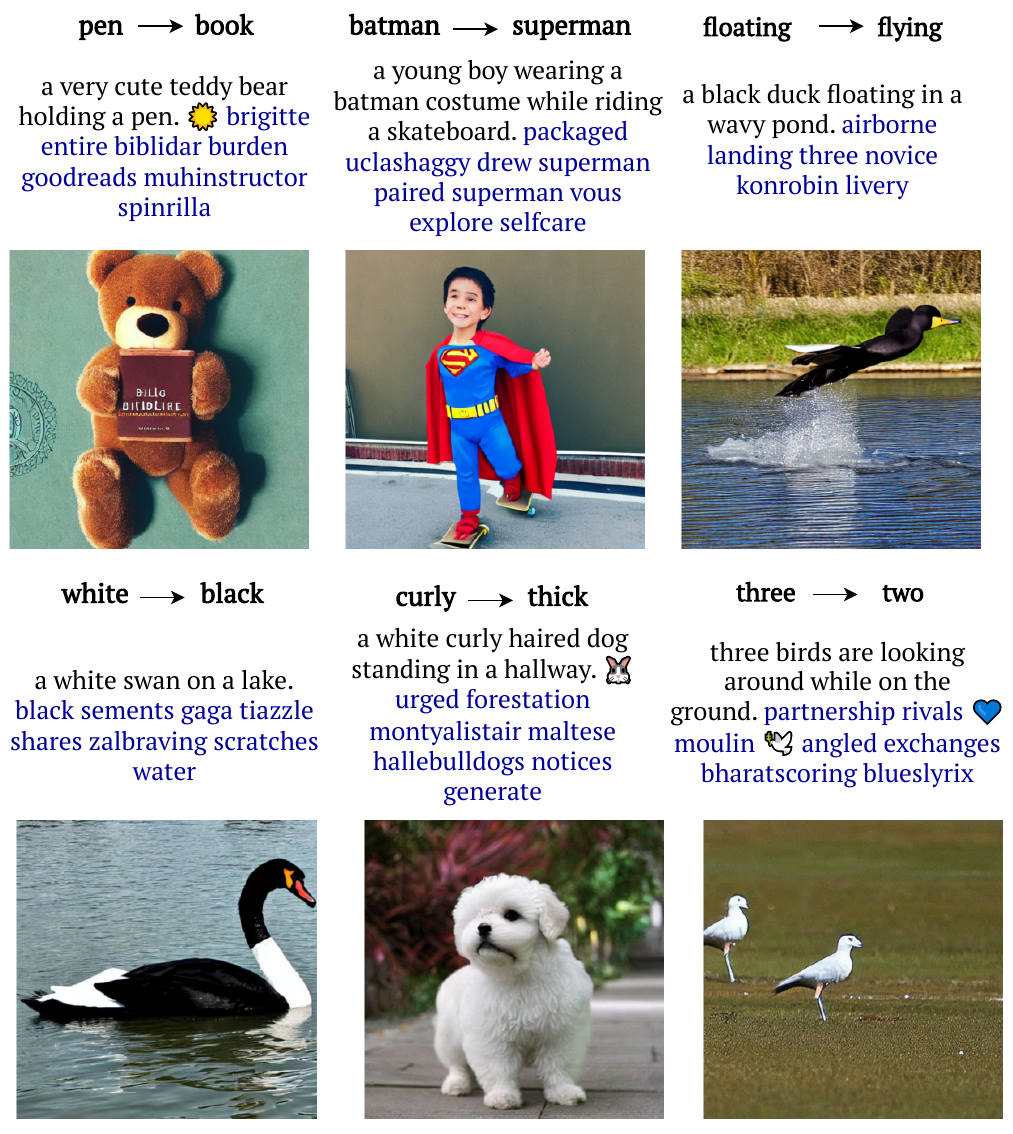}
  \caption{Some examples of successful attack on Stable Diffusion v1-4.}
  \label{fig:attack-transfer}
\end{figure}

\begin{figure}[h!]
  \centering
  \includegraphics[scale=0.37]{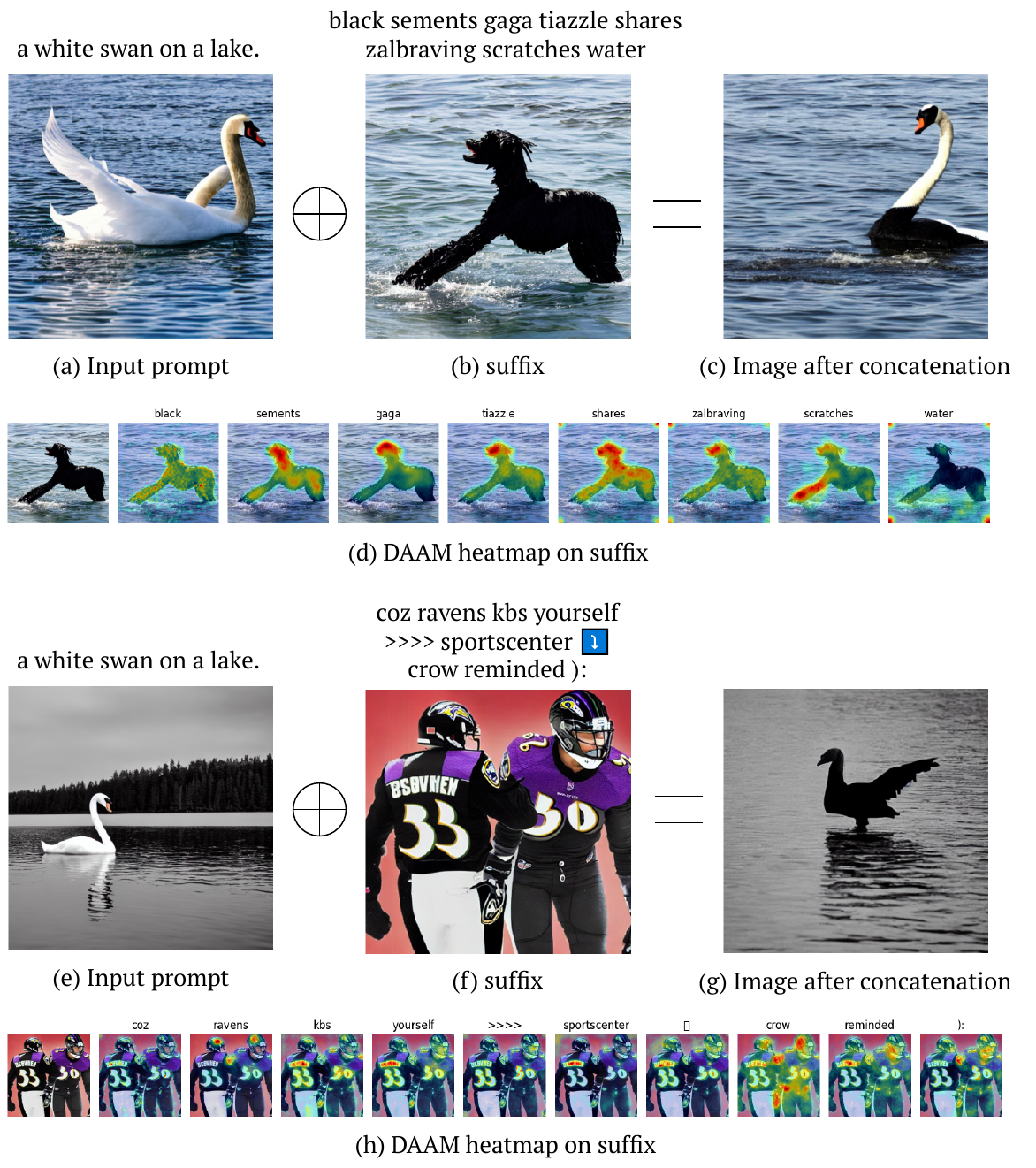}
  \caption{Examples demonstrating how the adversarial suffix independently dictates the SD model's output.}
  \label{fig:adjective-analysis-2}
\end{figure}

\section{Attack Transferability}
We employed Stable Diffusion v1-5, utilizing CLIP ViT-L/14 \cite{radford2021learning} as the pre-trained text encoder, for both unrestricted and restricted attack methods. Our investigation reveals that adversarial suffixes generated with this version of SD are ineffective when applied to Stable Diffusion v2-1, which uses a different pre-trained text encoder, OpenCLIP-ViT/H \cite{cherti2023reproducible}. However, we note that the attacks successfully transfer to Stable Diffusion v1-4. Examples generated using SD v1-4 are presented in Figure \ref{fig:attack-transfer}.

\section{Examples of Suffix Transferability} \label{app:E}
We provide a successful attack example in Figure \ref{fig:adjective-analysis-2} where the input and target prompts are \textit{``a white swan on a lake.''} and \textit{``a black swan on a lake.''} respectively. Figure \ref{fig:adjective-analysis-2}(a - d) represents the unrestricted setting. Figure \ref{fig:adjective-analysis-2}(a) and \ref{fig:adjective-analysis-2}(b) represent the images generated by the SD model using the text embedding of the input prompt and the suffix respectively while \ref{fig:adjective-analysis-2}(c) is the final image generated using the concatenated text embedding. As the suffix tokens are largely nonsensical, we employed DAAM \cite{tang-etal-2023-daam} to generate word attribution heatmap in order to delineate which suffix words correspond to which portion of the suffix image, as shown in Figure \ref{fig:adjective-analysis-2}(d). Similarly, we repeat the procedure for the restricted setting in Figure \ref{fig:adjective-analysis-2}(e - g). 
% Since the adversarial suffix can independently steer the SD model toward generating the desired output, we explored the potential for suffix transferability across different examples within the same POS tag. 
\begin{figure}[h]
  \centering
  \includegraphics[scale=0.55]{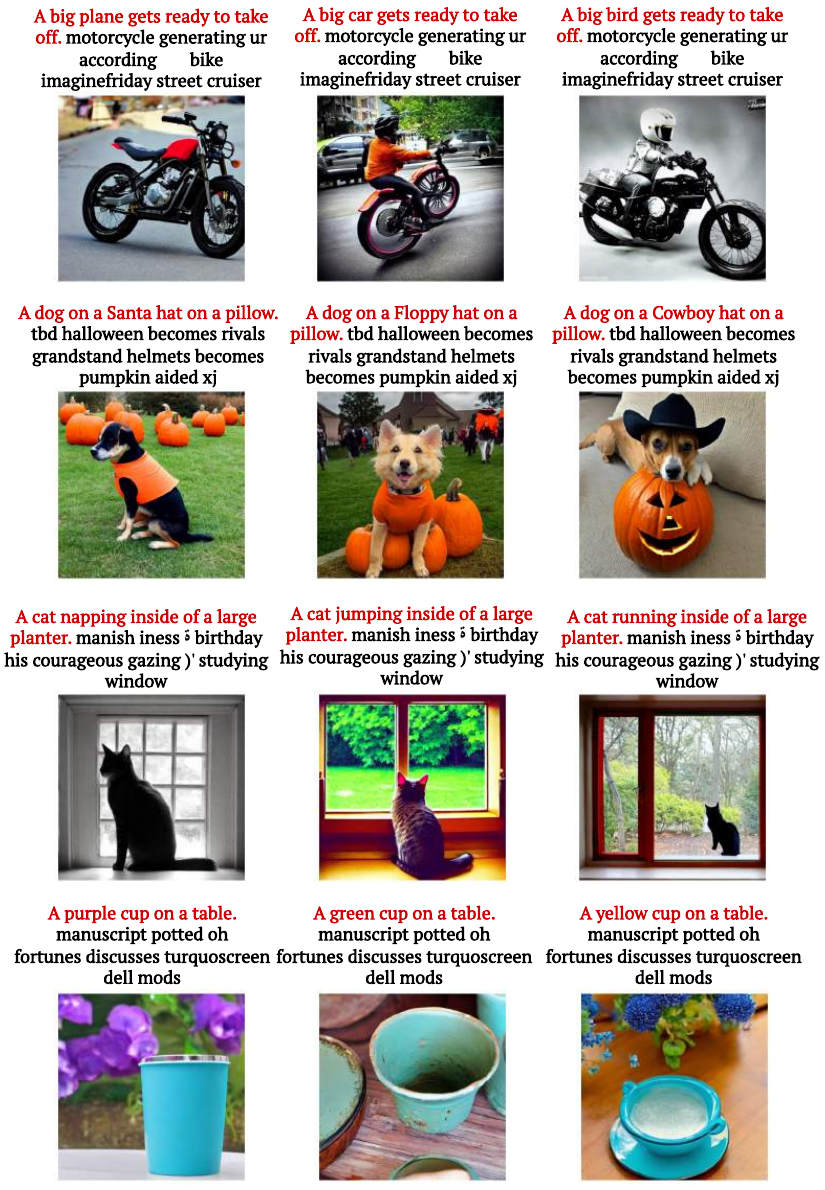}
  \caption{Examples of adversarial suffix transferability. The top two rows are the examples of \textit{noun} and \textit{proper noun} POS tags in unrestricted settings where the target words are ``motorcycle'' and ``Halloween'' respectively. The last two rows correspond to the examples of \textit{verb} and \textit{adjective} POS tags in restricted settings where the target words are ``watching'' and ``blue'' respectively.}
  \label{fig:suffix-transfer}
\end{figure}
\onecolumn
\section{More Examples of Adversarial Attack}
\begin{figure}[h!]
\centering
  \includegraphics[width=\textwidth, height=640px]{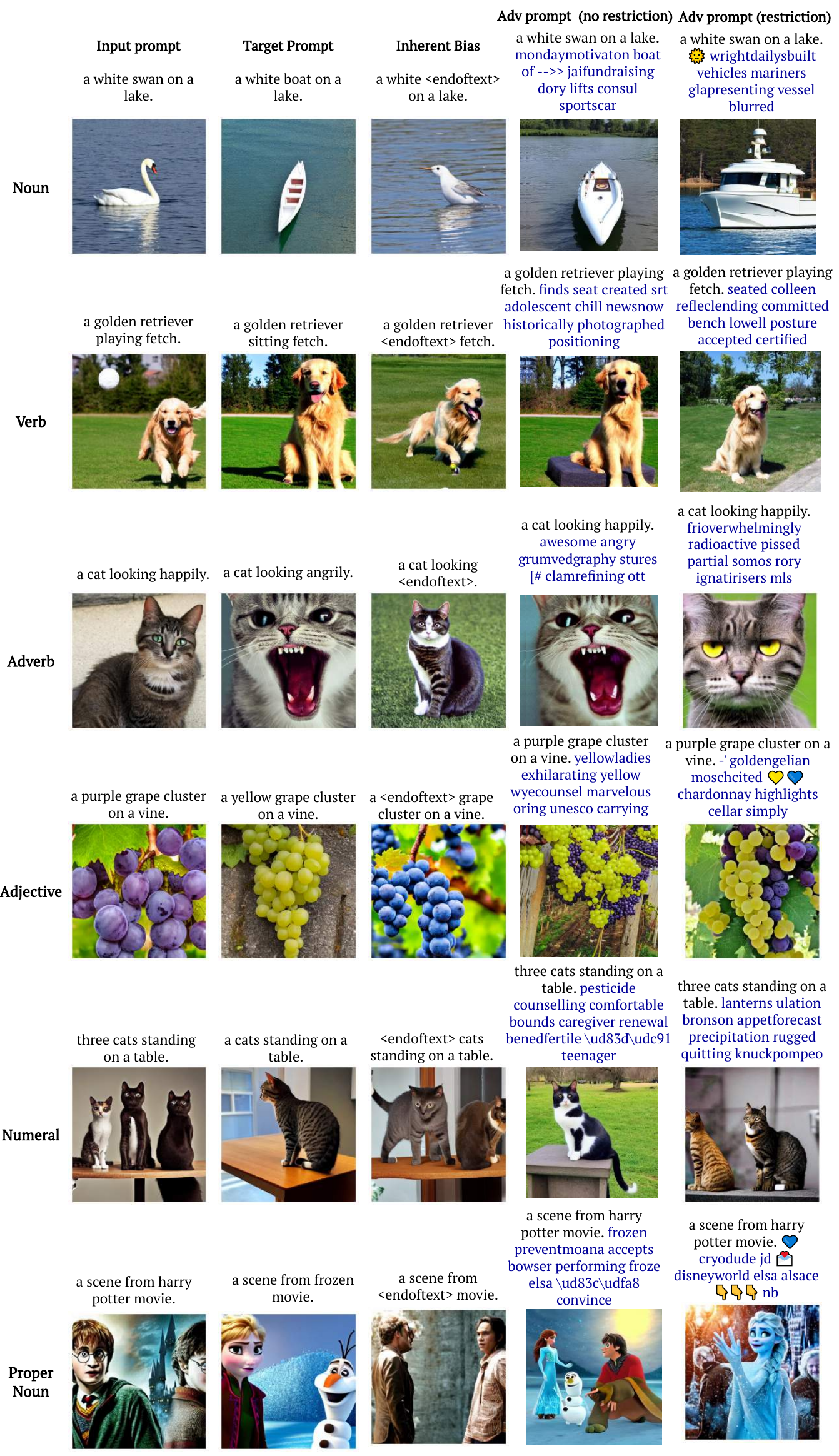}
  \caption{Few examples of successful attacks on out-of-dataset instances for each POS tag.}
  \label{fig:attack-example}
\end{figure}
\pagebreak

\onecolumn
\section{Some Examples of Adjective Color Fusion}\label{app:G}
\begin{figure}[h!]
\centering
  \includegraphics[width=\textwidth, height=640px]{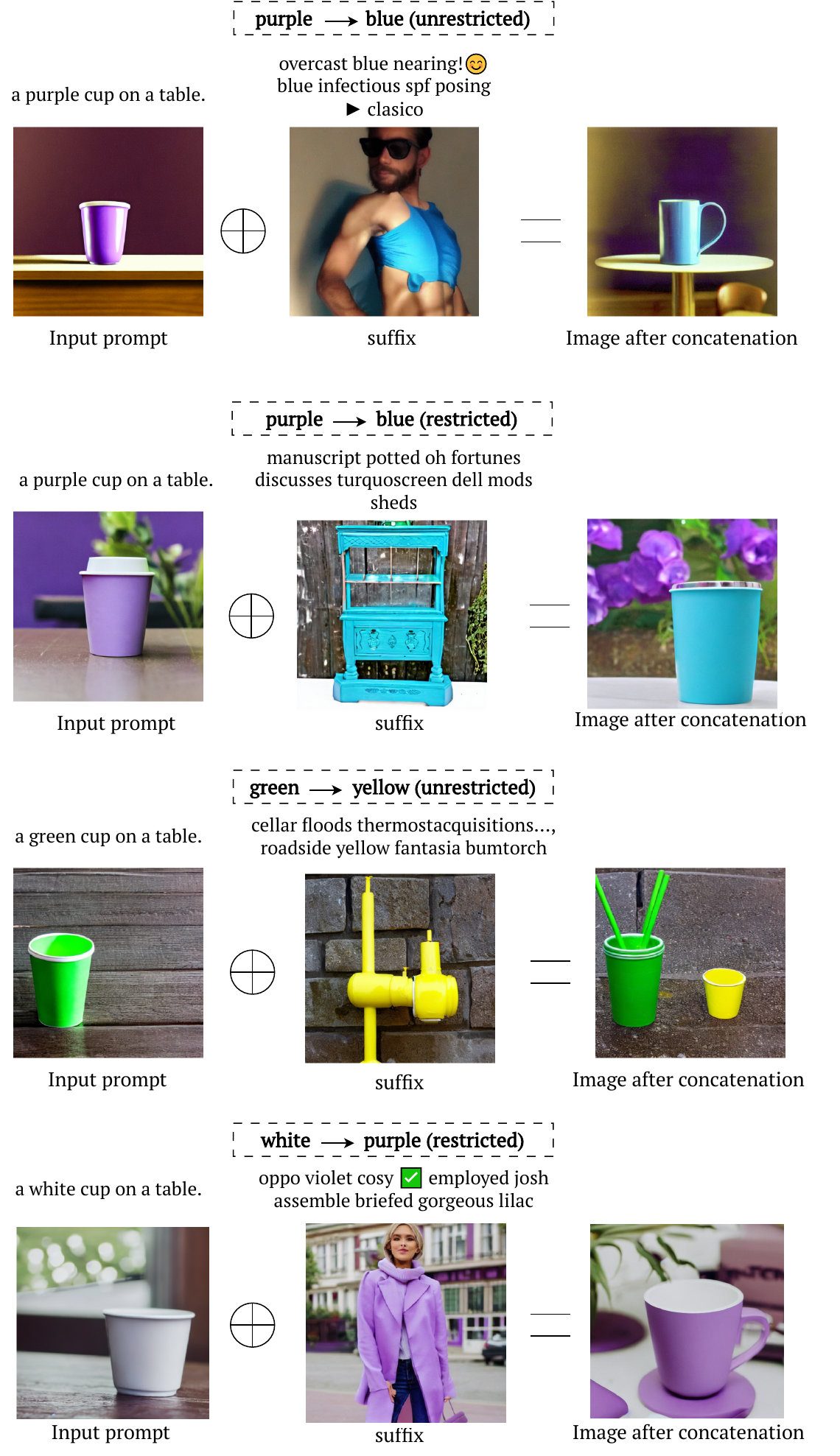}
  \caption{Examples of adversarial suffix found by the adversarial attack responsible for fusion of color adjectives.}
  \label{fig:adjective-entanglement}
\end{figure}
\pagebreak

\onecolumn
\section{Human Evaluation Template}\label{app:K}
\begin{figure}[h!]
\centering
  \includegraphics[width=\textwidth, height=640px]{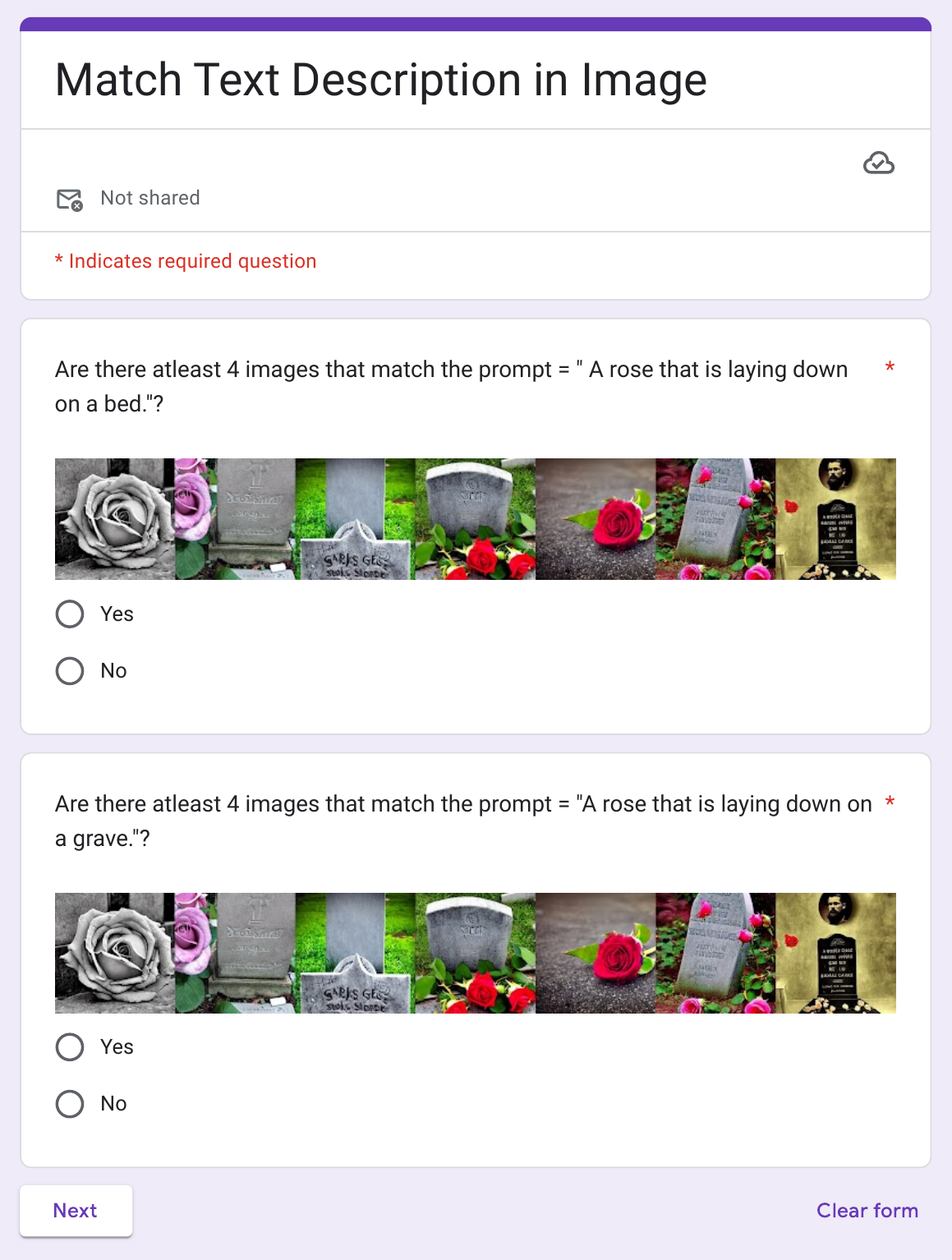}
  \caption{An overview of the human evaluation template.}
  \label{fig:eval-form}
\end{figure}
\pagebreak

%Too many small details may overwhelm the system.

\end{document}